\documentclass[10pt,twocolumn,letterpaper]{article}

\usepackage{cvm}
\usepackage{times}
\usepackage{epsfig}
\usepackage{graphicx}
\usepackage{amsmath}
\usepackage{amssymb}
\usepackage{adjustbox}

\usepackage{algorithmic}
\usepackage{algorithm}
\usepackage{booktabs}
\usepackage{multirow}
\usepackage[table]{xcolor}
\newcommand{\ours}{MT-PCR}

\usepackage[pagebackref=true,breaklinks=true,letterpaper=true,colorlinks,bookmarks=false]{hyperref}

\usepackage{cleveref}

\newcommand{\keywords}[1]{{\bf \emph{Keywords: #1}}}

\cvmfinalcopy 

\def\cvmPaperID{373} 

\ifcvmfinal\fi
\begin{document}

\title{MT-PCR: Hybrid Mamba-Transformer Network with Spatial Serialization for Point Cloud Registration}

\author{Bingxi Liu$^{1, 3, }$\thanks{Equal contribution.}, An Liu$^{2, *}$, Hao Chen$^{4}$, Huaqi Tao$^{1}$, Jinqiang Cui$^{3}$, Yiqun Wang$^{2, }$\thanks{Corresponding authors.}, Hong Zhang$^{1, \dag}$ 
\\ $^1$\small Southern University of Science and Technology, Shenzhen, China
\\ $^2$\small Chongqing University, Chongqing, China
\\ $^3$\small Pengcheng Laboratory, Shenzhen, China
\\ $^4$\small University of Cambridge, Cambridge, United Kingdom
}


\maketitle

\begin{abstract}
Point cloud registration (PCR) is a fundamental task in 3D computer vision and robotics. Most learning-based PCR methods rely on Transformer architectures, which suffer from quadratic computational complexity. This limitation restricts the resolution of point clouds that can be processed, inevitably leading to information loss. In contrast, Mamba, a recently proposed model based on state-space models, achieves linear computational complexity while maintaining strong long-range contextual modeling capabilities. However, directly applying Mamba to PCR tasks yields suboptimal performance due to the unordered and irregular nature of point cloud data. To address these challenges, we propose MT-PCR, the first point cloud registration framework that integrates Mamba and Transformer modules. Specifically, we serialize point cloud features using Z-order space-filling curves to enforce spatial locality, enabling Mamba to better model the geometric structure of the inputs. Additionally, we remove the order-indicator module commonly used in Mamba-based sequence modeling, leading to improved performance in our setting. The serialized features are then processed by an optimized Mamba encoder, followed by a Transformer-based feature refinement stage. Extensive experiments on multiple benchmarks demonstrate that MT-PCR outperforms Transformer-based and other state-of-the-art methods in both accuracy and efficiency, significantly reducing GPU memory usage and FLOPs.
\end{abstract}


\section{Introduction}

Point cloud registration (PCR) is a foundational task in 3D computer vision \cite{wang2019deep} and robotics \cite{yang2020teaser}, widely applied in domains such as 3D measurement \cite{sgu}, simultaneous localization and mapping (SLAM) \cite{cadena2016past}, augmented reality (AR) \cite{carmigniani2011augmented}, and autonomous driving \cite{lu2019l3}. PCR aims to estimate an optimal rigid transformation that aligns partially overlapping 3D point cloud pairs into a unified coordinate system. Despite significant advances, achieving efficient and accurate PCR remains challenging due to complex spatial structures \cite{wang20243dpcp}, partial overlaps \cite{guo2024low}, measurement noise, and large-scale scenes \cite{lu2023hregnet}.

\begin{figure}[t]
\centering
\includegraphics[width=1\linewidth]{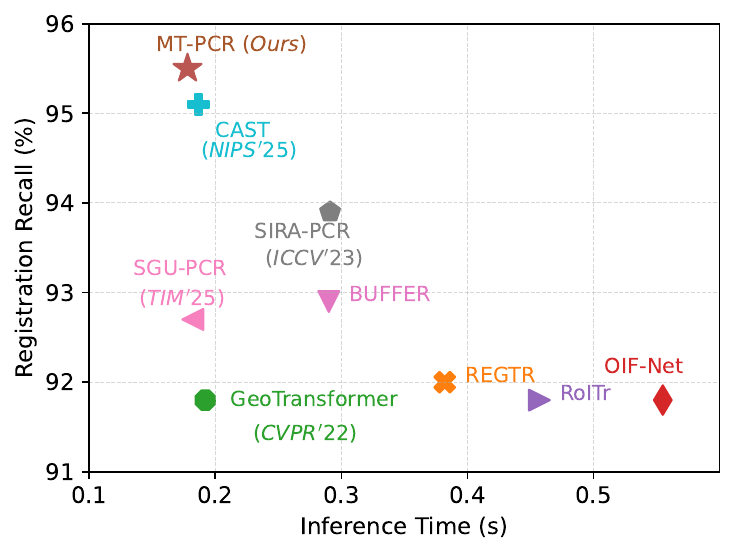}
\caption{\textbf{Registration recall and inference time comparison on 3DMatch.} Our method, MT-PCR, achieves the best registration performance while maintaining competitive inference efficiency, outperforming recent state-of-the-art methods such as CAST (NIPS’25) and SGU-PCR (TIM'25).
}
\label{fig:teaser}
\end{figure}

Recently, Transformer-based methods have demonstrated remarkable performance improvements in PCR tasks by leveraging powerful self-attention and cross-attention mechanisms \cite{qin2023geotransformer}. Transformers \cite{vaswani2017attention} effectively model global spatial relationships and feature correspondences across point clouds, significantly outperforming hand-crafted or convolution-based approaches. However, Transformers inherently exhibit quadratic computational complexity with respect to sequence length \cite{chou2024metala}, severely limiting their scalability and resolution capacity. Consequently, existing Transformer-based methods typically rely on downsampling point clouds to reduce computational overhead, which inevitably leads to the loss of important geometric information critical for precise alignment.

On the other hand, linear attention models, particularly the recently introduced Mamba model \cite{mamba}, have demonstrated promising results in efficiently capturing the long-range contextual dependencies in sequential modeling tasks. Unlike Transformers, Mamba leverages linear-complexity state-space architectures to approximate global context, thereby significantly enhancing computational efficiency and scalability for long sequences. Nevertheless, directly applying the Mamba model to PCR tasks leads to suboptimal registration accuracy, primarily due to the absence of explicit spatial modeling and inadequate handling of unstructured point cloud data.

Motivated by these observations, we propose MT-PCR, \textit{the first} hierarchical point cloud registration framework that combines the efficient sequence modeling capabilities of Mamba with the bidirectional spatial awareness enabled by cross-attention modules. To effectively adapt Mamba to irregular point cloud data, we introduce a feature serialization strategy based on Z-order space-filling curves, which enforces spatial locality and enhances compatibility between point cloud structures and Mamba's sequence modeling paradigm. Furthermore, we observe that removing the order-indicator tokens, which are typically employed in Mamba for sequential tasks, improves registration performance on 3D data.

\begin{figure}[t]
	\centering
	\includegraphics[width=1\linewidth]{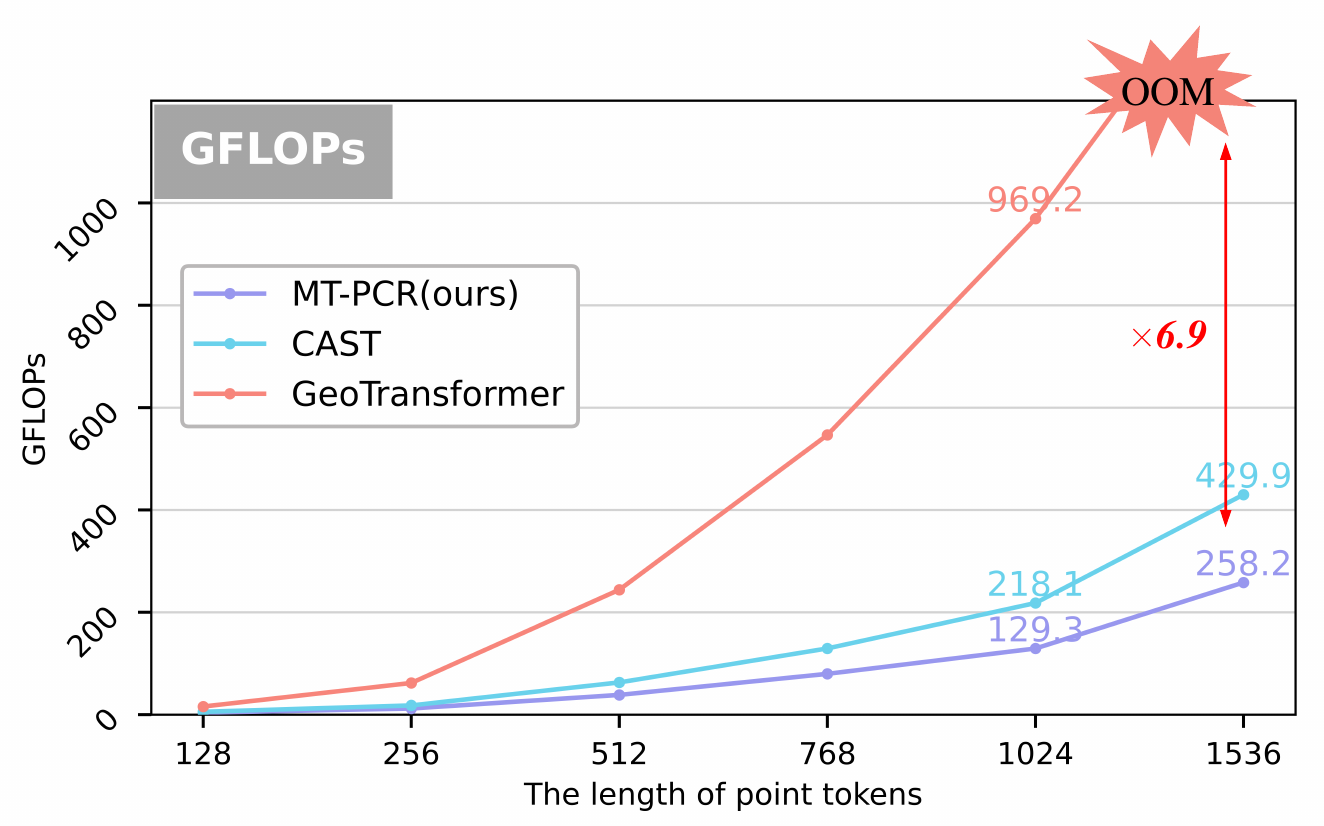}
	\caption{\textbf{FLOPs comparison under varying point token lengths.} MT-PCR scales significantly better than GeoTransformer and CAST, maintaining low computational overhead even as the input size increases. Notably, GeoTransformer suffers from \textit{out-of-memory} (OOM) issues at large resolutions, while MT-PCR remains efficient, achieving up to $6.9 \times$ lower FLOPs at 1536 tokens.
	}
	\label{fig:teaser_2}
\end{figure}

Extensive experiments on standard PCR benchmarks, including the widely used 3DMatch~\cite{zeng20173dmatch}, 3DLoMatch~\cite{huang2021predator}, KITTI~\cite{geiger2012kitti}, and ETH-Challenges~\cite{Pomerleau2012eth} datasets, demonstrate that MT-PCR significantly outperforms existing Transformer-based methods and other state-of-the-art (SOTA) approaches, as illustrated in Fig.~\ref{fig:teaser}. Moreover, our method achieves these improvements while drastically reducing GPU memory usage and computational overhead in terms of FLOPs, thereby validating the scalability and efficiency of our hybrid architecture, as shown in Fig.~\ref{fig:teaser_2}.

In summary, \textit{our contributions} encompass four key aspects as follows:

\begin{itemize}
	\item We introduce MT-PCR, the \textit{\textbf{first}} hybrid Mamba-Transformer framework for hierarchical point cloud registration, which leverages linear-complexity sequence modeling alongside bidirectional cross-attention mechanisms.
	
	\item We propose a spatially aware serialization method based on Z-order space-filling curves, enabling Mamba to effectively process unstructured point cloud data in a sequential format.
	
	\item We demonstrate that the order-indicator module, commonly adopted in Mamba-based sequence modeling, is unnecessary in our context, and its removal results in improved performance on PCR tasks.
	
	\item We conduct comprehensive experiments across multiple benchmarks, showing that MT-PCR achieves state-of-the-art performance while significantly enhancing computational efficiency in terms of memory usage and FLOPs.
\end{itemize}

\section{Related Work}

\subsection{Point Cloud Registration}

PCR aims to estimate a rigid transformation between point clouds. Early methods primarily adopted hand-crafted descriptors to represent local features, typically leveraging local geometric attributes~\cite{salti2014shot} or spatial distribution histograms~\cite{rusu2009fpfh,rusu2008pfh}. Nonetheless, the representational capacity of hand-crafted features is inherently limited, often leading to matching failures in noisy or complex scenarios.

Recently, learning-based 3D descriptors have demonstrated significant advantages. PerfectMatch~\cite{gojcic2019perfect} employs a Smoothed Density Value representation to learn discriminative features. PPF~\cite{deng2018ppfnet} extracts globally context-aware, patch-wise features using a PointNet~\cite{qi2017pointnet}-based architecture. FCGF~\cite{choy2019fcgf} utilizes a sparse 3D convolutional encoder--decoder network for dense descriptor learning. SpinNet~\cite{ao2021spinnet} proposes a 3D cylindrical convolutional network with specialized coordinate-system designs to achieve rotation invariance. Predator~\cite{huang2021predator} integrates graph convolutions with cross-attention mechanisms, explicitly optimizing overlap-region prediction for low-overlap scenarios.

Inspired by Transformer architectures, recent studies have incorporated attention mechanisms into PCR networks to enhance robustness. CoFiNet~\cite{yu2021cofinet} pioneers the integration of self-attention and cross-attention modules for coarse-level feature matching, combined with optimal transport theory to refine fine-grained correspondences. GeoTransformer~\cite{qin2023geotransformer} introduces geometry-enhanced self-attention and a local-to-global registration framework to ensure pose-estimation consistency. RoITr~\cite{yu2023roitr} develops a rotation-invariant Transformer architecture based on point-pair features, strengthening the robustness of coarse-to-fine frameworks. DiffusionPCR~\cite{chen2023diffusionpcr} incorporates diffusion models to enable iterative refinement of feature matching. CAST~\cite{huang2024consistency} integrates consistency-aware mechanisms and point-guided strategies to suppress interference from irrelevant regions while enhancing feature matching capabilities. However, these methods are inherently limited by the quadratic computational complexity $\mathcal{O}(N^2)$ of Transformer architectures, which restricts the resolvable point-cloud size; consequently, downsampling high-density point clouds inevitably discards geometric detail.

\subsection{State Space Models and Mamba}

State Space Models (SSMs)~\cite{kalman1960new} originate from control theory and have garnered attention as efficient alternatives to Transformers for sequence modeling. Classical SSMs such as S4~\cite{gu2022parameterization} demonstrate the ability to model long-range dependencies with linear complexity by leveraging diagonal-plus-low-rank state transitions and HiPPO initialization. These models often suffer from computational inefficiencies or limited scalability.

To address these issues, Mamba~\cite{mamba} introduces a selective SSM mechanism, where input-conditioned parameterization enables selective information flow. By combining input-dependent state updates with hardware-aware parallelization, Mamba achieves strong performance while maintaining linear-time inference. Variants such as Vision Mamba~\cite{vim} and Vmamba~\cite{liu2024vmamba} adapt Mamba to image-level tasks through bidirectional and cross-selective scanning strategies, showcasing its potential as a backbone for visual understanding.

Recent works attempt to transfer the benefits of SSMs and Mamba to 3D point cloud data. MetaLA~\cite{chou2024metala} integrates Mamba into a meta-sequential attention framework, emphasizing lightweight modeling for point cloud perception tasks. PointMamba~\cite{liu2024point, liang2024pointmamba} employs an octree-based serialization scheme to impose order on unordered point cloud data, enabling causal dependencies compatible with Mamba’s design. Mamba3D~\cite{han2024mamba3d} extends this line of research by incorporating Mamba into a 3D-aware architecture. It enhances Mamba’s spatial awareness through local geometry integration and various pretraining strategies, demonstrating improved scalability and representational capacity on 3D datasets. However, Mamba3D targets object-level recognition and does not explicitly address the unique demands of PCR. In contrast, our work proposes the first Mamba-based architecture specifically tailored for PCR. We also address key challenges in unordered modeling and build a hierarchical framework that combines Mamba’s global modeling capabilities with local attention and cross-scale refinement.

\section{Problem Definition}

Given two partially overlapping 3D point clouds, $X = \{\mathbf{x}_i \in \mathbb{R}^3 \mid i = 1, 2, \dots, M\}$ and $Y = \{\mathbf{y}_j \in \mathbb{R}^3 \mid j = 1, 2, \dots, N\}$, which are denoted as the source and target, respectively, the goal of PCR is to estimate an optimal rigid transformation that aligns $X$ with $Y$. This transformation is characterized by a rotation matrix $\mathbf{R} \in \mathrm{SO}(3)$ and a translation vector $\mathbf{t} \in \mathbb{R}^3$. The solution is typically obtained by minimizing a weighted sum of squared distances between corresponding point pairs in a predicted correspondence set $C$:
\begin{equation}
\min_{\mathbf{R}, \mathbf{t}} \sum_{(\mathbf{x}_k, \mathbf{y}_k) \in C} w_k \left\|\mathbf{R}\mathbf{x}_k + \mathbf{t} - \mathbf{y}_k\right\|_2^2,
\end{equation}
where $w_k$ denotes the weight associated with each correspondence $(\mathbf{x}_k, \mathbf{y}_k)$.

\section{Method}

In this section, we first introduce background on Transformers and Mamba in Sec.~\ref{sec:pre}. We then provide an overview of our framework, MT-PCR, in Sec.~\ref{sec:overview}. The spatial serialization strategy is described in detail in Sec.~\ref{z_order}, followed by the design of the Mamba encoder in Sec.~\ref{encoder}. Finally, we introduce the loss functions used for training in Sec.~\ref{loss}.

\begin{figure*}[t]
\centering
\includegraphics[width=\linewidth]{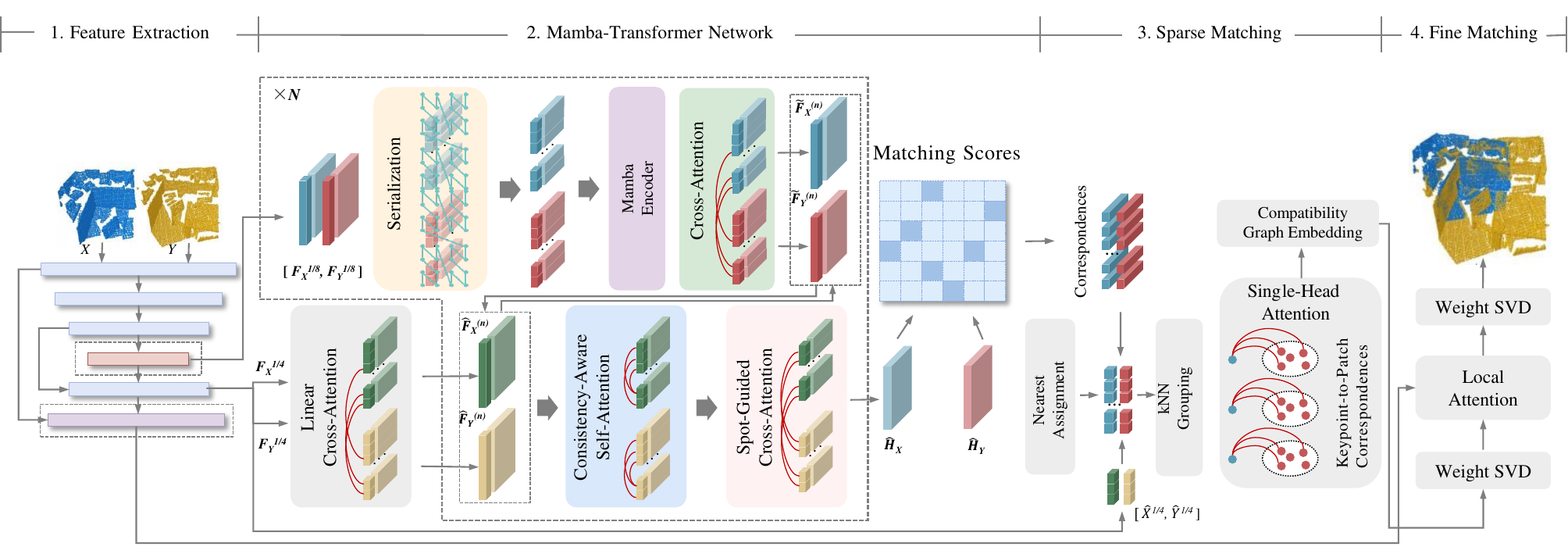}
\caption{\textbf{Overview of the MT-PCR Framework.} The proposed pipeline consists of four stages: multi-scale feature extraction, coarse matching, sparse correspondence refinement, and fine registration. Notably, the coarse matching stage incorporates Mamba encoders with spatial serialization to model global geometric context efficiently.
}
\label{fig:pipeline}
\end{figure*}

\subsection{Preliminaries}
\label{sec:pre}

\textbf{Attention Mechanisms and Transformer.}
Transformers, constructed by stacking self-attention and cross-attention modules, have been widely employed in PCR tasks to effectively model global dependencies and feature correspondences. The self-attention mechanism computes attention weights within the same point set to capture internal feature interactions, while cross-attention identifies correspondences between two distinct point sets. Formally, given queries $\mathbf{Q}$, keys $\mathbf{K}$, and values $\mathbf{V}$, attention is computed as follows:
\begin{equation}
\text{Attention}(\mathbf{Q}, \mathbf{K}, \mathbf{V}) = \text{softmax}\left(\frac{\mathbf{Q}\mathbf{K}^{\top}}{\sqrt{d_k}}\right)\mathbf{V},
\end{equation}
where $d_k$ denotes the dimensionality of key vectors. Despite their effectiveness, Transformers suffer from quadratic computational complexity with respect to sequence length, limiting their scalability to larger point clouds.

\textbf{State Space Models and Mamba.}
SSMs can be viewed as linear time-invariant, multi-input multi-output (MIMO) systems. Mathematically, a continuous-time SSM is described by a set of ordinary differential equations (ODEs):
\begin{equation}
\mathbf{h}'(t) = \mathbf{A}\mathbf{h}(t) + \mathbf{B}\mathbf{x}(t), 
\end{equation}
\begin{equation}
\quad \mathbf{y}(t) = \mathbf{C}\mathbf{h}(t) + \mathbf{D}\mathbf{x}(t),
\end{equation}
where $\mathbf{x}(t) \in \mathbb{R}^L$, $\mathbf{h}(t)\in \mathbb{R}^N$, $\mathbf{y}(t) \in \mathbb{R}^L$, and $\mathbf{h}'(t)\in \mathbb{R}^N$ represent the continuous-time input, hidden state, output, and the derivative of the hidden state, respectively. $\mathbf{A} \in \mathbb{R}^{N \times N}$ is the state matrix, $\mathbf{B} \in \mathbb{R}^{N \times L}$ is the input matrix, $\mathbf{C} \in \mathbb{R}^{L \times N}$ is the output matrix, and $\mathbf{D} \in \mathbb{R}^{L \times L}$ is the feed-through matrix.

The continuous-time system can be discretized into a discrete-time SSM via zero-order hold (ZOH) discretization. The parameters $\mathbf{A}$ and $\mathbf{B}$ of the discrete-time SSM can be obtained by introducing the sampling step $\Delta$ and applying a Taylor expansion. In this case, the parameters can be approximated as:
\begin{equation}
\mathbf{\overline{A}} = \exp(\Delta\mathbf{A}), 
\quad \mathbf{\overline{B}} = \left(\exp(\Delta\mathbf{A}) - \mathbf{I}\right) (\Delta\mathbf{A})^{-1} \cdot \Delta\mathbf{B}, 
\end{equation}
which results in the discrete form:
\begin{equation}
\mathbf{h}_k = \mathbf{\overline{A}}\mathbf{h}_{k-1} + \mathbf{\overline{B}}\mathbf{x}_k, 
\end{equation}
\begin{equation}
\quad \mathbf{y}_k = \mathbf{\overline{C}}\mathbf{h}_k + \mathbf{\overline{D}}\mathbf{x}_k,
\end{equation}
where $\mathbf{x}_k$, $\mathbf{h}_k$, and $\mathbf{y}_k$ represent discrete-time input, state, and output vectors, respectively.

Inspired by the above formulations, Gu and Dao proposed the Mamba model~\cite{mamba}, a novel variant of SSMs that introduces input-dependent and time-varying system parameters. Specifically, $\Delta$, $\mathbf{A}$, and $\mathbf{B}$ dynamically adapt according to the input $\mathbf{x}_t$. The Mamba model achieves sequence-modeling performance comparable to Transformers while maintaining linear computational complexity during inference. Although direct parallel computation is challenging, this issue is addressed via a global convolutional expansion:
\begin{equation}
\mathbf{K} = (\mathbf{C}\mathbf{\overline{B}}, \mathbf{C}\mathbf{\overline{A}}\mathbf{\overline{B}}, \dots, \mathbf{C}\mathbf{\overline{A}}^{M-1}\mathbf{\overline{B}}), \quad \mathbf{y} = \mathbf{x} \ast \mathbf{K},
\end{equation}
where $M$ denotes the input sequence length and $\mathbf{K}$ represents the global convolution kernel. This convolutional formulation significantly improves computational efficiency. Further details can be found in the Mamba framework~\cite{mamba}.

\subsection{\ours~Overview}
\label{sec:overview}

As illustrated in Fig.~\ref{fig:pipeline}, \ours~employs a coarse-to-fine, multi-level feature matching architecture to achieve accurate PCR. It consists of the following key steps:

\textbf{Multi-scale Feature Extraction:} Kernel Point Convolution (KPConv) is utilized to encode input point clouds into multi-scale feature representations spanning from the original dense point cloud (level $L_0$) to coarser downsampled point clouds (level $L_k$). The decoder-generated feature maps are denoted as $\mathbf{F}^{1/k} = \{\mathbf{F}^{1/k}_{X}, \mathbf{F}^{1/k}_{Y}\}$, corresponding to nodes $X^{1/k}$ and $Y^{1/k}$ downsampled from the original point clouds $X$ and $Y$, respectively. The key points at the topmost downsampling level are termed superpoints, serving as anchor points for subsequent feature matching.

\textbf{Mamba-Transformer Coarse Matching:} Stacked Mamba feature extraction modules are applied at the superpoint level to construct correspondence sets $\mathcal{C}_s = \{(x_i, y_j)\}$ within overlapping local regions. Based on these correspondences, a differentiable matching matrix $\mathbf{M} \in \mathbb{R}^{N_s \times N_s}$ is constructed via compatibility graph convolutional networks, where each matrix element $m_{ij}$ indicates the matching confidence between superpoint pairs $(x_i, y_j)$. 

\textbf{Sparse Correspondence Refinement:} Based on coarse matching results, discriminative keypoints are identified within semi-dense node neighborhoods denoted as ${X}^{1/4}$ and ${Y}^{1/4}$. Virtual correspondences for these keypoints are then predicted using a lightweight attention module. To further eliminate spatially inconsistent outliers, a compatibility graph embedding network is employed, yielding a set of high-confidence inliers $\mathcal{I}_{\text{inlier}}$.

\textbf{Fine Registration.} An alignment between the source $\mathbf{X}^{1/2}$ and the target $\mathbf{Y}^{1/2}$ is performed first using the initial transformation $(\hat{\mathbf{R}}_0, \hat{\mathbf{t}}_0)$. Then, a lightweight local attention module is applied to the point sets to estimate refined dense correspondences. These correspondences are then used to compute the final rigid transformation $(\hat{\mathbf{R}}, \hat{\mathbf{t}})$ via weighted SVD.

\noindent \textbf{Challenges and Insights.} Replacing the self-attention module with the Mamba module offers a promising pathway toward efficient global modeling due to its linear-time complexity. However, this substitution poses a fundamental challenge: \textbf{point clouds are inherently unordered and spatially irregular}, while Mamba is designed for causal, structured sequences as found in natural language processing. Unlike Transformers, which are permutation-invariant and better suited for unordered inputs, Mamba requires well-structured sequential inputs to operate effectively. This discrepancy raises a critical question: \textit{How can we convert 3D point cloud data into meaningful one-dimensional sequences suitable for Mamba while preserving spatial coherence and geometric information?}

To overcome these challenges, we introduce a traversal-based serialization strategy, along with its variants, to generate multiple point-cloud sequences. This approach maximizes the retention of topological associations through multi-path spatial traversal. Additionally, we observe that removing order-indicator tokens, which are typically used in Mamba for sequential tasks, improves registration performance.


\subsection{Z-order-based Spatial Serialization}
\label{z_order}

The superpoint features $\mathbf{F}^{1/k}$ obtained from KP-Conv are inherently unordered, which poses challenges for state space models like Mamba that rely on directional, sequential processing. To enable effective sequence modeling, we convert the 3D point cloud into a one-dimensional sequence while preserving spatial locality.

\begin{algorithm}[!htpb]
	\caption{Z-order Encoding for 3D Point Clouds}
	\label{alg:zorder}
	\begin{algorithmic}[1]
		\REQUIRE Grid coordinates $\mathbf{G} = \{(x_i, y_i, z_i)\}_{i=1}^N$, depth $d$, batch labels $\mathbf{b}$
		\ENSURE Z-order serialization code $\pi(\mathbf{p}_i)$ for each point
		
		\FORALL {$\mathbf{p}_i = (x_i, y_i, z_i) \in \mathbf{G}$}
		\STATE Convert $x_i, y_i, z_i$ to $d$-bit binary integers
		\STATE Interleave bits from $(x_i, y_i, z_i)$ to get Morton code $m_i$
		\IF{batch labels $\mathbf{b}$ are provided}
		\STATE Concatenate batch ID as prefix: 
		\STATE \quad $m_i \leftarrow (\text{batch}_i \ll 3d) \mid m_i$
		\ENDIF
		\ENDFOR
		\STATE \RETURN Z-order code list $\{\pi(\mathbf{p}_i) = m_i\}_{i=1}^N$
	\end{algorithmic}
\end{algorithm}

Formally, given a point cloud $\mathcal{X} = \{\mathbf{x}_i \mid \mathbf{x}_i \in \mathbb{R}^3, i = 1, \dots, N\}$, we define a bijective serialization function $f: \mathbb{R}^3 \rightarrow \mathbb{R}$ that maps each point to a scalar index: $f: \mathbf{x}_i \mapsto s_i, \quad s_i \in \mathbb{R}, \quad \forall \mathbf{x}_i \in \mathcal{X}.$

Among various strategies, space-filling curves (SFCs) like the Z-order curve (Morton code) are particularly effective, as they preserve spatial locality:
\begin{equation}
\| \mathbf{x}_i - \mathbf{x}_j \|_2 \approx 0 \Rightarrow |s_i - s_j| \approx 0.
\end{equation}
This ensures that neighboring points in 3D space remain adjacent in the serialized 1D sequence, which is critical for maintaining structural coherence in Mamba-based modeling.

As shown in Alg.~\ref{alg:zorder}, we adopt Z-order serialization by first quantizing 3D coordinates to a discrete grid and then interleaving the bits of the $d$-bit integer representations of $(x, y, z)$ to compute Morton codes. The resulting Z-order indices $\pi(\mathbf{p}_i)$ define a spatially-aware serialization path, enabling Mamba to effectively process the point cloud sequence while preserving geometric proximity.

\subsection{\textbf{Mamba Encoder}}
\label{encoder}

\begin{figure}[t]
\begin{center}
\includegraphics[width=\linewidth]{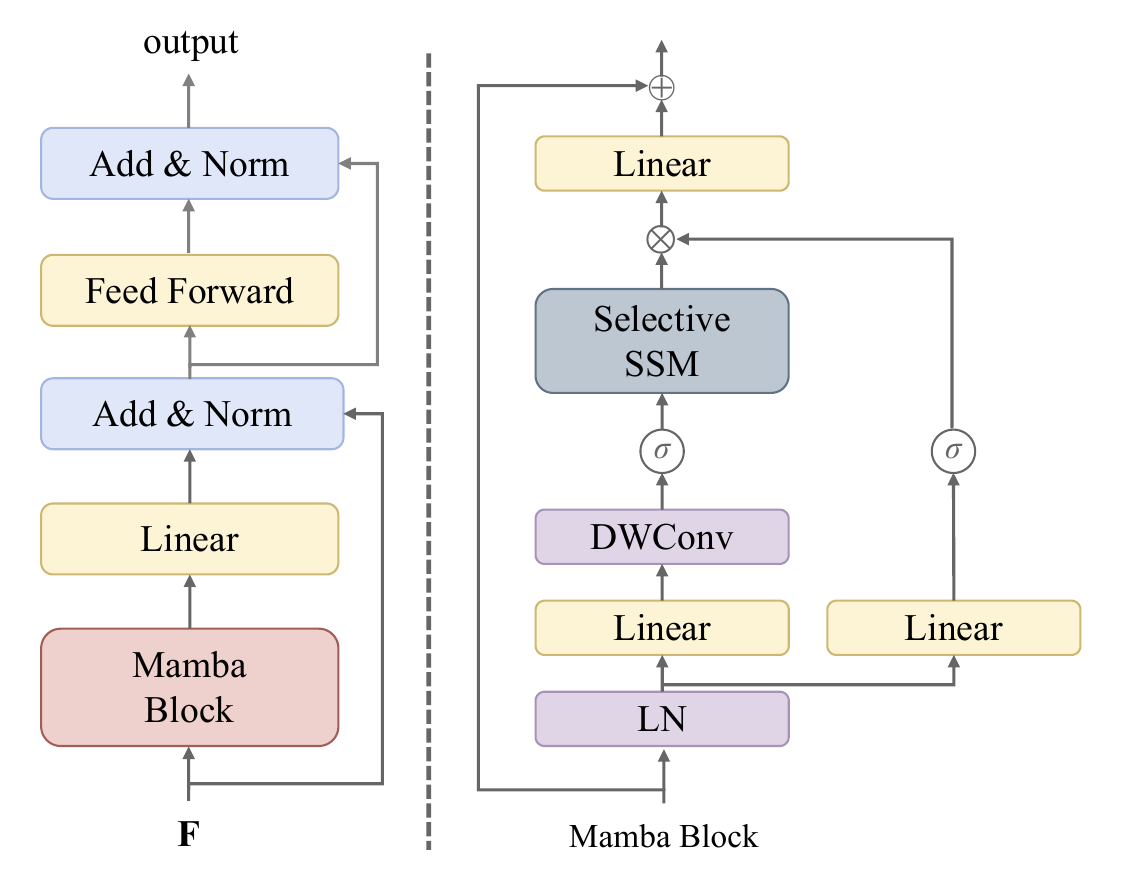}
\caption{\textbf{Architecture of the Mamba Encoder and Block.} The left diagram illustrates the Mamba Encoder with residual connections and feed-forward networks (FNNs). The right diagram shows the internal structure of a Mamba Block, which centers around the SelectiveSSM.}
\label{fig:mamba}
\end{center}
\end{figure}

{\noindent
After serialization, the feature tokens $z$ are fed into hybrid Mamba--Transformer
blocks to extract hierarchical geometric features. Specifically, features from the two inputs are processed sequentially by a Mamba block to capture global context and a Transformer block to refine correspondences. Each Mamba block consists of Layer Normalization (LN), a Selective State Space Model (SelectiveSSM), depth-wise
separable convolutions (DW), and residual connections. The architecture is
illustrated in Fig.~\ref{fig:mamba}, and the forward pass of the $l$-th block is
given by:
}

\begin{equation}
    F'_{l-1} = \text{LN}(F_{l-1}),
\end{equation}
\begin{equation}
    F'_l = \sigma\left(\text{DW}\left(\text{Linear}\left(F'_{l-1}\right)\right)\right),
\end{equation}
\begin{equation}
    F''_l = \sigma\left(\text{Linear}\left(F'_{l-1}\right)\right),
\end{equation}
\begin{equation}
    F_l = \text{Linear}\left(\text{SelectiveSSM}(F'_l) \odot F''_l\right) + F_{l-1},
\end{equation}
where $F_l \in \mathbb{R}^{2n \times C}$ denotes the output features, and $\sigma$ represents the SiLU activation function. The SelectiveSSM forms the core of the Mamba block, adaptively modeling contextual dependencies through input-conditioned parameterization. We adopt a minimalist yet effective design~\cite{liang2024pointmamba} that complies with Occam’s razor, simplifying the implementation while retaining its expressiveness.

\subsection{Loss Functions}
\label{loss}
Our overall loss function supervises four key modules in the hierarchical registration framework: \textit{keypoint detection}, \textit{coarse matching}, \textit{keypoint matching}, and \textit{dense registration}. Each sub-objective is addressed with a tailored loss component to guide the network effectively at different stages of alignment.

\noindent \textbf{Keypoint Detection.} We supervise keypoint detection~\cite{li2019usip} with $\mathcal{L}_p$. This loss encourages spatial alignment between predicted keypoints from the source and target point clouds while accounting for uncertainty. Specifically, we define:
\begin{equation}
\begin{aligned}
\mathcal{L}_p &= \frac{1}{N} \sum_{i=1}^N \left( \log \tilde{\sigma}_i + \frac{\|\mathbf{x}_i - \mathbf{y}_{j^*(i)}\|}{\tilde{\sigma}_i} \right) \\
&\quad + \frac{1}{M} \sum_{j=1}^M \left( \log \tilde{\sigma}_j + \frac{\|\mathbf{y}_j - \mathbf{x}_{i^*(j)}\|}{\tilde{\sigma}_j} \right),
\end{aligned}
\end{equation}
where $\mathbf{x}_i \in \mathbb{R}^3$ and $\mathbf{y}_j \in \mathbb{R}^3$ are keypoints from the reference and transformed point clouds, respectively. The indices $j^*(i) = \arg\min_j \|\mathbf{x}_i - \mathbf{y}_j\|$ and $i^*(j) = \arg\min_i \|\mathbf{y}_j - \mathbf{x}_i\|$ denote nearest neighbors. The uncertainty-aware weighting term $\tilde{\sigma}$ is computed as the average predicted variance from both matched keypoints:
\begin{equation}
\tilde{\sigma}_i = \frac{1}{2} \left( \sigma_{x,i} + \sigma_{y,j^*(i)} \right), \quad 
\tilde{\sigma}_j = \frac{1}{2} \left( \sigma_{y,j} + \sigma_{x,i^*(j)} \right).
\end{equation}

\noindent \textbf{Coarse Matching.} We utilize two weighted cross-entropy losses: the spot-matching loss $\mathcal{L}_s$ for layer-wise coarse scores $\mathbf{P}^{(l)}$ and the coarse-matching loss $\mathcal{L}_c$ for final coarse scores $\mathbf{P}$:
\begin{equation}
\mathcal{L}_s = -\frac{1}{L}\sum_{l=1}^L \frac{1}{\sum_{(i,j)\in \mathcal{C}}o_{ij}} \sum_{(i,j)\in \mathcal{C}} o_{ij}\log \mathbf{P}_{ij}^{(l)},
\end{equation}
\begin{equation}
\begin{aligned}
\mathcal{L}_c = & -\frac{1}{\sum_{(i,j)\in \mathcal{C}}o_{ij}} \sum_{(i,j)\in \mathcal{C}} o_{ij}\log \mathbf{P}_{ij}\\
&-\frac{1}{|\mathcal{N}_X|}\sum_{k\in \mathcal{N}_X} \log(1-\hat{o}_k^X)\\
&-\frac{1}{|\mathcal{N}_Y|}\sum_{k\in \mathcal{N}_Y} \log(1-\hat{o}_k^Y),
\end{aligned}
\end{equation}
where $\mathcal{N}_X$ and $\mathcal{N}_Y$ denote sets of semi-dense nodes in the source and target point clouds without correspondences. The overlap ratio $o_{ij}$ measures how much two local patches overlap. Further details can be found in CAST~\cite{huang2024consistency}.

\noindent \textbf{Keypoint Matching.} We employ three losses to supervise similarity estimation, correspondence prediction, and consistency filtering. First,
we use the InfoNCE loss~\cite{oord2018infonce}, $\mathcal{L}_f$, to maximize the similarity between descriptors $d_x$ and $d_{p_x}$ of true correspondences $(x,p_x)$ and to minimize the similarity between descriptors $d_x$ and $d_{n_x}$ of false correspondences $(x,n_x)$:
\begin{equation}
\mathcal{L}_f = -\mathbb{E}\left[ \log\frac{e{(d_x^{\mathsf{T}}Wd_{p_x})}}{e^{(d_x^{\mathsf{T}}Wd_{p_x}}) + \sum_{n_x\in N_x}e{(d_x^{\mathsf{T}}Wd_{n_x})}}\right],
\end{equation}
where $W$ is a symmetric learnable weight matrix. Next, we use an $\ell_2$ loss, $\mathcal{L}_k$, to supervise predicted correspondences $\hat{\mathbf{y}}$ by minimizing:
\begin{equation}
\mathcal{L}_k = \mathbb{E}_{(x, \hat{y})} \left\| \mathbf{R}x + \mathbf{t} - \hat{y} \right\|_2.
\end{equation}

Finally, for consistency filtering, we define binary ground-truth labels based on whether a correspondence is an inlier (i.e., its distance is below the threshold $R_f$) and supervise inlier confidence using binary cross-entropy:
\begin{equation}
\mathcal{L}_i = \text{BCE}(\text{score}, \text{inlier label}).
\end{equation}

\noindent \textbf{Dense Registration.} The dense registration module is supervised using translation and rotation losses:
\begin{equation}
\mathcal{L}_t = \left\|\hat{\mathbf{t}} - \mathbf{t}\right\|_2, \quad \mathcal{L}_R = \left\|\hat{\mathbf{R}}^{\mathsf{T}}\mathbf{R} - \mathbf{I}\right\|_F.
\end{equation}
The final training loss is formulated as:
\begin{equation}
\begin{aligned}
    \mathcal{L} = &\mathcal{L}_p + \lambda_s \mathcal{L}_s + \lambda_c \mathcal{L}_c + \lambda_f \mathcal{L}_f + \lambda_k \mathcal{L}_k + \lambda_i \mathcal{L}_i \\
    &+ \lambda_t \mathcal{L}_t + \lambda_R \mathcal{L}_R,
\end{aligned}
\end{equation}
where $\lambda_s, \lambda_c, \lambda_f, \lambda_k, \lambda_i, \lambda_t, \lambda_R$ are hyperparameters used to balance the contribution of each term by aligning their numerical scales.

\section{Evaluation}
\label{sec:evaluation}

\newcommand{\first}[1]{\textcolor[HTML]{FE0000}{\textbf{#1}}}
\newcommand{\second}[1]{\textcolor[HTML]{3531FF}{\textbf{#1}}}
\newcommand{\third}[1]{\textcolor[HTML]{1f821c}{\textbf{#1}}}

\begin{table*}[!htpb]
\small 
\caption{Evaluation results on indoor RGBD point cloud datasets. \first{First}, \second{second}, and \third{third} best results are color-highlighted.}
\label{table:3dmatch}
\centering
\begin{tabular}{cc|c|ccccc|ccccc}
\toprule
& Dataset & Source & \multicolumn{5}{c|}{3DMatch} & \multicolumn{5}{c}{3DLoMatch}\\
\cmidrule{1-13}
& & & \multicolumn{5}{c|}{Registration Recall (\%)} & \multicolumn{5}{c}{Registration Recall (\%)} \\
& Samples & & 5000 & 2500 & 1000 & 500 & 250 & 5000 & 2500 & 1000 & 500 & 250 \\
\cmidrule{1-13}
\multirow{6}{*}{\rotatebox[origin=c]{90}{Descriptor-based}} & PerfectMatch~\cite{gojcic2019perfect}
& \textit{CVPR'19} & 78.4 & 76.2 & 71.4 & 67.6 & 50.8
& 33.0 & 29.0 & 23.3 & 17.0 & 11.0 \\
& FCGF~\cite{choy2019fcgf} & \textit{ICCV'19}
& 85.1 & 84.7 & 83.3 & 81.6 & 71.4
& 40.1 & 41.7 & 38.2 & 35.4 & 26.8 \\
& D3Feat~\cite{bai2020d3feat} & \textit{CVPR'20}
& 81.6 & 84.5 & 83.4 & 82.4 & 77.9
& 37.2 & 42.7 & 46.9 & 43.8 & 39.1 \\
& SpinNet~\cite{ao2021spinnet} & \textit{CVPR'21}
& 88.6 & 86.6 & 85.5 & 83.5 & 70.2
& 59.8 & 54.9 & 48.3 & 39.8 & 26.8 \\
& YOHO~\cite{wang2022yoho} & \textit{MM'22}
& 90.8 & 90.3 & 89.1 & 88.6 & 84.5
& 65.2 & 65.5 & 63.2 & 56.5 & 48.0 \\
& Predator~\cite{huang2021predator} & \textit{CVPR'21}
& 89.0 & 89.9 & 90.6 & 88.5 & 86.6
& 59.8 & 61.2 & 62.4 & 60.8 & 58.1 \\
\cmidrule{1-13}
\multirow{10}{*}{\rotatebox[origin=c]{90}{Correspondence-based}}
& REGTR~\cite{yew2022regtr} & \textit{CVPR'22} & & & 92.0 & &
& & & 64.8 & & \\
& GeoTransformer~\cite{qin2023geotransformer} & \textit{TPAMI'23}
& 92.0 & 91.8 & 91.8 & 91.4 & 91.2
& 75.0 & 74.8 & 74.2 & 74.1 & 73.5 \\
& OIF-Net~\cite{yang2022one} & \textit{NIPS'22}
& 92.4 & 91.9 & 91.8 & 92.1 & 91.2 
& 76.1 & 75.4 & 75.1 & 74.4 & 73.6 \\
& RoITr~\cite{yu2023roitr} & \textit{CVPR'23}
& 91.9 & 91.7 & 91.8 & 91.4 & 91.0 
& 74.7 & 74.8 & 74.8 & 74.2 & 73.6 \\
& PEAL~\cite{yu2023peal} & \textit{CVPR'23}
& 94.4 & 94.1 & 94.1 & 93.9 & 93.4 
& \second{79.2} & 79.0 & 78.8 & 78.5 & 77.9 \\
& BUFFER~\cite{ao2023buffer} & \textit{CVPR'23} & & & 92.9 & & 
& & & 71.8 & & \\
& SIRA-PCR~\cite{chen2023sira}
& \textit{ICCV'23} & 93.6 & 93.9 & 93.9 & 92.7 & 92.4 
& 73.5 & 73.9 & 73.0 & 73.4 & 71.1\\
& DiffusionPCR~\cite{chen2023diffusionpcr}
& \textit{arXiv'24} & 94.4 & 94.3 & \third{94.5} & 94.0 & 93.9 
& {80.0} & \first{80.4} & {79.2} & {78.8} & 78.8 \\
& CAST ~\cite{huang2024consistency} & \textit{NIPS'25} & & & \second{95.2} & &  & & & 75.1 & & \\
& SGU-PCR ~\cite{huang2024consistency} & \textit{TIM'25} & 93.4 & 92.8 & 92.7 & 92.4 & 91.5 & 75.4 & 74.9 & 74.3 & 74.1 & 74.0 \\
\rowcolor{gray!15}& \ours & \textit{Ours}& & & \first{95.5} & &  & & & \third{75.4} & & \\
\bottomrule
\end{tabular}
 \vspace{-6pt}
\end{table*}

In this section, we conduct extensive experiments to evaluate the performance of our proposed \ours~on both indoor RGB-D datasets (3DMatch~\cite{zeng20173dmatch} and 3DLoMatch~\cite{huang2021predator}) and the outdoor 3D LiDAR dataset (KITTI~\cite{geiger2012kitti}). To further validate the generalization of the proposed method, we also test it on a 2D LiDAR dataset (ETH-Challenging~\cite{Chall_3d}).

\subsection{Implementation Details}

We employ the AdamW~\cite{loshchilov2018adamw} optimizer, an initial learning rate of $1 \times 10^{-4}$, and a weight decay of $1 \times 10^{-4}$. A stepwise learning rate scheduler is adopted, decaying the learning rate by 10\% every 5 training steps. Gradient clipping with a threshold of 0.5 is used to stabilize the training process. Except for the experiments in Tab.~\ref{table:length_of_token}, which are performed on an NVIDIA A800 GPU, the rest of the experiments are conducted on a single RTX 3090 GPU. The number of training epochs is set to 5 for 3DMatch and 40 for KITTI. For the KITTI and nuScenes datasets, we set: $\lambda_f = \lambda_i = 1, \lambda_r = 20, \lambda_t = 5, \lambda_s = 0.1, \lambda_c = 0.2, \lambda_k = 1.$ For the 3DMatch dataset, we use: $\lambda_c = 1, \lambda_k = 10$.

\subsection{Indoor Scenes: 3DMatch \& 3DLoMatch}

As shown in Tab.~\ref{table:3dmatch}, we evaluate MT-PCR alongside several SoTA methods on the widely used indoor PCR benchmarks 3DMatch~\cite{zeng20173dmatch} and 3DLoMatch~\cite{huang2021predator}, which respectively represent high-overlap ($>$30\%) and low-overlap (10\%–30\%) scenarios. To ensure fair and consistent comparison of runtime and resource consumption, all methods are re-implemented within a unified PyTorch framework and tested under identical hardware conditions. Our benchmarks include descriptor-based feature matching approaches as well as non-iterative correspondence-based registration methods.

Since MT-PCR relies on sparse keypoint matching, it typically detects around 1000 keypoints. Therefore, we report the Registration Recall (RR) based on 1000 correspondences, aligning with prior work~\cite{huang2024consistency}. Notably, despite using only 1000 correspondences, our method outperforms competing dense matching approaches that rely on 1000 or more sampled points. This demonstrates the superior effectiveness of our sparse matching strategy.

On the 3DMatch dataset, our method achieves a new state-of-the-art RR of 95.5\%. On the more challenging 3DLoMatch dataset, MT-PCR attains 75.4\% RR, outperforming all descriptor-based baselines and non-iterative matching methods. Thanks to the linear attention mechanism enabled by Mamba, our approach also demonstrates superior efficiency, achieving the fastest runtime among all compared methods.

While our RR on 3DLoMatch is slightly lower than that of PEAL~\cite{yu2023peal} and DiffusionPCR~\cite{chen2023diffusionpcr}, these methods incur over \textit{10$\times$} the runtime overhead compared to ours. This favorable efficiency–accuracy trade-off is particularly important for real-world applications with limited computational resources.

\subsection{Outdoor Scenes: KITTI Odometry}

Tab.~\ref{tab:kitti} summarizes the evaluation of MT-PCR on the KITTI Odometry dataset~\cite{geiger2012kitti}, a standard benchmark for autonomous driving scenarios. We follow the conventional split: sequences 00–05 for training, 06–07 for validation, and 08–10 for testing. To ensure high-quality training samples, point cloud pairs with at least 10 meters spatial separation are selected. Ground-truth transformations are obtained by aligning GPS/IMU trajectories refined via ICP.

We evaluate MT-PCR using Relative Rotation Error (RRE), Relative Translation Error (RTE), and Registration Recall (RR). Our comparisons include SoTA descriptor-based methods such as FCGF~\cite{choy2019fcgf}, D3Feat~\cite{bai2020d3feat}, SpinNet~\cite{ao2021spinnet}, Predator~\cite{huang2021predator}, and correspondence-based approaches including CoFiNet~\cite{yu2021cofinet}, GeoTransformer~\cite{qin2023geotransformer}, OIF-Net~\cite{yang2022one}, PEAL~\cite{yu2023peal}, DiffusionPCR~\cite{chen2023diffusionpcr}, MAC~\cite{zhang2023mac}, and CAST~\cite{huang2024consistency}.

MT-PCR achieves SoTA results in both RR and RRE, outperforming the previous best model (DiffusionPCR). Notably, this is \textbf{\textit{the first time}} RRE has dropped below 0.2 degrees. This highlights MT-PCR's strong ability to predict accurate rotations. Although its RTE shows slight differences compared to other leading methods, it remains highly competitive and consistent, underscoring robustness in outdoor LiDAR registration.

\begin{table}[!ht]
\small
\caption{Registration performance on KITTI dataset. \first{First}, \second{second}, and \third{third} best results are color-highlighted.}
\centering
\begin{tabular}{lccc}
\toprule
Model & RTE (cm) & RRE (°) & RR (\%) \\
\cmidrule{1-4}
3DFeat-Net~\cite{yew20183dfeat} & 25.9 & 0.25 & 96.0 \\
FCGF~\cite{choy2019fcgf} & 9.5 & 0.30 & 96.6 \\
D3Feat~\cite{bai2020d3feat} & 7.2 & 0.30 & 99.8 \\
SpinNet~\cite{ao2021spinnet} & 9.9 & 0.47 & 99.1 \\
Predator~\cite{huang2021predator} & 6.8 & 0.27 & 99.8 \\
CoFiNet~\cite{yu2021cofinet} & 8.2 & 0.41 & 99.8 \\
GeoTrans~\cite{qin2023geotransformer} & 6.8 & 0.24 & 99.8 \\
OIF-Net~\cite{yang2022one} & 6.5 & \second{0.23} & 99.8 \\
PEAL~\cite{yu2023peal} & 6.8 & \second{0.23} & 99.8 \\
DiffusionPCR~\cite{chen2023diffusionpcr} & \third{6.3} & \second{0.23} & 99.8 \\
MAC~\cite{zhang2023mac} & 8.5 & 0.40 & 99.5 \\
CAST~\cite{huang2024consistency} & \first{2.5} & \third{0.27} & \first{100.0} \\
\rowcolor{gray!15} \ours  & \second{2.6} & \first{0.16} & \first{100.0} \\
\bottomrule
\end{tabular}
\label{tab:kitti}
\end{table}

\subsection{Efficiency Study}


\begin{table}[!t]
\small
\centering
\caption{\textbf{Efficiency Comparison at Varying Token Lengths.} Computational efficiency for different methods across increasing token lengths. * On an 80GB GPU, GeoTransformer measured a GFLOPs of 1803 and memory usage of 24,614 MB at token length 1536.}
\label{table:length_of_token}
\resizebox{\linewidth}{!}{
\begin{tabular}{l|cccccc}
\toprule
Length of Token & 128 & 256 & 512 & 768 & 1024 & 1536 \\
\midrule
& \multicolumn{6}{c}{\emph{GFLOPs} $\downarrow$} \\
\midrule
GeoTrans ~\cite{qin2023geotransformer} & 16 & 62 & 244 & 547 & 969 & out of memory* \\
CoFiNet ~\cite{yu2021cofinet} & 8 & 18 & 58 & 147 & 479 & out of memory* \\
CAST ~\cite{huang2024consistency} & 6  & 18  & 63 & 129 & 218  & 430 \\
MT-PCR (\emph{ours}) & \textbf{4}  & \textbf{12}  & \textbf{39} & \textbf{80} & \textbf{129}  & \textbf{258} \\
\midrule
& \multicolumn{6}{c}{\emph{GPU Memory} (MB) $\downarrow$} \\
\midrule
GeoTrans ~\cite{qin2023geotransformer} & 176 & 634 & 2463 & 5510 & 12335 & out of memory* \\
CoFiNet ~\cite{yu2021cofinet} & 147 & 573 & 1947 & 3709 & 8732 & out of memory* \\
CAST ~\cite{huang2024consistency} & 120 & \textbf{334} & 1185 & 2491 & 4189 & 11028 \\
MT-PCR (\emph{ours}) & \textbf{119} & 335 & \textbf{1162} & \textbf{2428} & \textbf{4091} & \textbf{10591} \\
\midrule
& \multicolumn{6}{c}{\emph{FPS}  $\uparrow$} \\
\midrule
GeoTrans ~\cite{qin2023geotransformer} & 5.97 & 5.43 & 4.79 & 3.27 & 2.83 & out of memory* \\
CoFiNet ~\cite{yu2021cofinet} & 5.44 & 4.68	& 3.35 & 2.75 & 1.68 & out of memory* \\
CAST ~\cite{huang2024consistency} & \textbf{6.52} & \textbf{6.19} & \textbf{5.27} & 4.50 & 3.69 & 2.63 \\
MT-PCR (\emph{ours}) & 6.17 & 5.92 & 5.22 & 4.59 & \textbf{4.02} & \textbf{3.00} \\
\bottomrule
\end{tabular}
}
\vspace{-3pt}
\end{table}

A core motivation of our work is to address the quadratic time complexity $\mathcal{O}(N^2)$ and high resource consumption inherent in Transformer-based PCRs. To evaluate the efficiency of our design, we conduct a scalability study by varying the serialized point cloud sequence length and measuring performance in terms of FLOPs and memory usage.

We adopt a progressive stress-testing strategy on a 24GB GPU, gradually increasing input sequence length until memory constraints are reached. As shown in Tab.~\ref{table:length_of_token}, our method achieves superior inference speed and lower GFLOPs across all lengths. Specifically, at a sequence length of 1536, our method reduces computation cost by 85.4\% (to $1/6.9$ GFLOPs) and memory usage by 58.3\% (to $1/2.4$) compared to a Transformer baseline—while maintaining comparable or better performance. This highlights the significant computational advantage of our model for long sequence modeling.

\subsection{Visualization}

\begin{figure}[t]
\begin{center}
\vspace{-4pt}
\includegraphics[width=\linewidth]{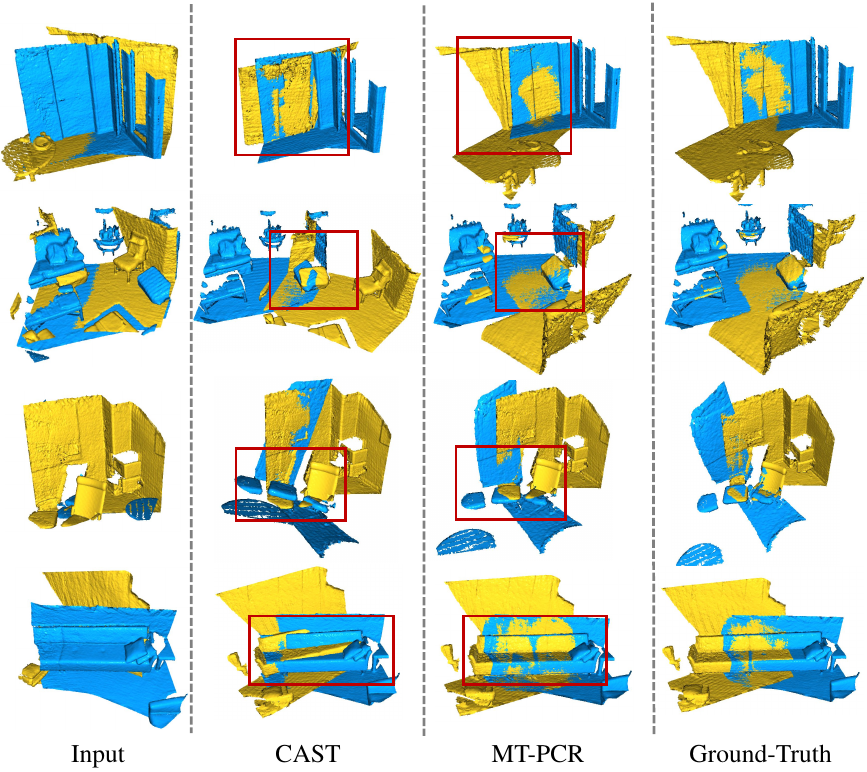}
\vspace{-10pt}
\caption{\textbf{Qualitative registration results} of CAST and MT-PCR compared with the ground truth alignment on 3DMatch dataset. We present four examples in four rows, which demonstrate the robustness and accuracy of our method.
}\label{fig:vis_3dmatch}
\end{center}
\end{figure}

Fig.~\ref{fig:vis_3dmatch} provides qualitative results on registration performance on 3DMatch. Compared to CAST, MT-PCR produces more accurate alignment results, especially in challenging regions highlighted by red boxes. The scenes registered by MT-PCR are visibly closer to the ground truth, demonstrating superior robustness and geometric consistency.

\subsection{Ablation Study}

\noindent \textbf{Effectiveness of Architecture.}  
The results in Tab.~\ref{tab:ablation_mamba_transformer} demonstrate the advantage of the hybrid Mamba-Transformer (HMT) architecture over single backbone models. HMT consistently outperforms both Mamba and Transformer alone in terms of RR, PMR, and PIR, confirming that the combination effectively leverages complementary strengths.

\begin{table}[ht]
	\centering
	\caption{\textbf{Comparison of different architectures.}}
	\label{tab:ablation_mamba_transformer}
	\begin{tabular}{l|ccc}
		\toprule
		Method & RR (\%) & PMR (\%) & PIR (\%)  \\ %
		\midrule
		Mamba & 92.34 & 94.17 & 71.68 \\
		Transformer & 94.96 & 95.88 & 75.54 \\
		HMT & \textbf{95.54} & \textbf{96.87} & \textbf{79.65} \\
		\bottomrule
	\end{tabular}
\end{table}

\begin{table}[t]
\small
\centering
\caption{\textbf{Ablation studies on linear attention methods.} The tested linear attention methods are incorporated within our proposed hybrid architecture and improvements. The method names are used here solely for ease of reference.}
\begin{tabular}{l|ccc}
\toprule
Model  & RR (\%) & PMR (\%) & PIR (\%)  \\ %
\midrule
\ Based~\cite{arora_simple_2024} & 94.75 & 96.33 & 78.24\\
\ Samba~\cite{ren2024samba} & 94.44 & 96.06 & 76.46\\
\ Metala~\cite{chou2024metala}  & 94.53 & 95.36 & 73.17 \\
\ Mamba & \textbf{95.54} & \textbf{96.87} & \textbf{79.65}\\
\bottomrule
\end{tabular}
\label{tab:arch}
\end{table}

\noindent \textbf{Effectiveness of Mamba.}
In addition to Mamba, several newly proposed models with linear computational complexity have recently emerged, as shown in Tab.~\ref{tab:arch}. To evaluate their effectiveness in the context of PCR, we conducted ablation studies comparing Mamba with representative linear attention models. The results suggest that the improved Mamba variant achieves the best performance among these linear models in registration accuracy.

\begin{table}[t]
\centering
\caption{\textbf{Ablation studies on serialization strategies.}}
\begin{tabular}{l|ccc}
        \toprule
        \textbf{Strategy}  & RR (\%) & PMR (\%) & PIR (\%)  \\ %
        \midrule
        baseline  & 93.47 & 96.24 & 76.84  \\
        \midrule
        $ \text{hilbert}\times 3 $ & 94.36 & 96.42 & 77.09  \\
        $ \text{hilbert}^{\mathsf{T}}\times 3 $ & 94.08 & 95.97 & 75.10 \\
        $ \text{hilbert}, \text{hilbert}^{\mathsf{T}}, \text{z} $  & 95.07 & 96.84 & 78.83 \\
        $ \text{xyz, zxy, yxz} $ & 94.00 & 96.33 & 76.57\\
        $ \text{z}^{\mathsf{T}}\times 3 $  & 94.37 & 96.51 & 78.16 \\
        $ \text{z}\times 3 $ & \textbf{95.54} & \textbf{96.87} & \textbf{79.65} \\    
        \bottomrule
        \end{tabular}
\label{tab:serialization}
\end{table}

\noindent {\textbf{Effectiveness of Serialization.}} We next examine the impact of different serialization strategies using spatial space-filling curves. In particular, we compare the performance of two representative curves: Hilbert and Z-order, along with their axis-permuted variants: Trans-Hilbert and Trans-Z-order. We also include the commonly used XYZ-order and its variant.

As shown in Tab.~\ref{tab:serialization}, methods using space-filling curve serialization outperform traditional baselines, with the Z-order strategy achieving the best overall performance. We attribute this to the locality-preserving nature of space-filling curves, which provide more logically ordered sequences for state space models. This continuous spatial scanning offers a natural inductive bias for Mamba to model geometry-aware sequences.

\begin{table}[t]
\small
\centering
\caption{\textbf{Ablation studies on order strategies.}}
\begin{tabular}{l|ccc}
\toprule
Setting  & RR (\%) & PMR (\%) & PIR (\%)  \\ %
\midrule
\ None & \textbf{95.54} & \textbf{96.87} & \textbf{79.65}\\
\ Order indicator & 94.74 & 96.42 & 77.75\\
\bottomrule
\end{tabular}
\label{tab:order}
\vspace{-3pt}
\end{table}

\noindent {\textbf{Effectiveness of the Order Indicator.}} In prior work such as PointMamba~\cite{liang2024pointmamba}, the Order-Indicator module was necessary because the model used two spatial curves (e.g., Hilbert and T-Hilbert) within the same Mamba encoder to perform bidirectional modeling. Without explicit order distinction, the mixed sequences would interfere with each other in the latent space, degrading performance. In contrast, our method performs all bidirectional information exchange in a subsequent cross-attention module, while the Mamba encoder operates only on a unidirectional sequence generated through dynamic serialization. This design preserves the geometric continuity of the point cloud from the beginning. Introducing an Order-Indicator in our case would artificially break spatial adjacency, weakening Mamba’s ability to capture local topology. As shown in Tab.~\ref{tab:order}, adding the Order-Indicator leads to a 0.8\% drop in RR, confirming its negative impact. Therefore, we choose to remove this module in order to maximize the preservation of geometric structure during Mamba processing.

\noindent
\textbf{Effectiveness of Mamba and Z-order on Rotation Estimation.}
As summarized in Tab.\ref{tab:zorder}, removing the Mamba encoder significantly
degrades RRE from 0.161$^\circ$ to 0.273$^\circ$, confirming its critical role
in capturing global geometric context beyond the backbone. Furthermore,
omitting Z-order serialization increases RRE to 0.174$^\circ$, demonstrating
that the spatial locality preserved by Z-order provides a superior inductive
bias for high-precision rotation estimation.

\noindent {\textbf{Sensitivity of Z-order Depth.}} We conducted a sensitivity analysis on the depth parameter ($d$) of the Z-order space-filling curve. As shown in Tab.~\ref{tab:zorder_depth}, increasing $d$ from 8 to 16 yields only a minor improvement in RR, indicating that the model's performance is not highly sensitive to this parameter.

\begin{table}[t]
\small
\centering
\caption{\textbf{Ablation Study on Rotation Estimation Components.}}
\begin{tabular}{l|ccc}
\toprule
Setting  & RR (\%) & RRE (°) & PTE (cm)  \\ %
\midrule
\ w/o zorder & \textbf{100.0} & 0.174 & 2.64\\
\ w/o mamba & \textbf{100.0} & 0.273 & 2.56\\
\ MT-PCR & \textbf{100.0} & \textbf{0.161} & \textbf{2.55} \\
\bottomrule
\end{tabular}
\label{tab:zorder}
\vspace{-3pt}
\end{table}




\begin{table}[ht]
\centering
\caption{\textbf{Sensitivity of Z-order depth parameter $d$ using RR (\%).}}
\label{tab:zorder_depth}
\begin{tabular}{c|ccc}
\toprule
Z-order Depth ($d$) & 8 & 12 & 16 \\
\midrule
RR (\%) & 94.90 & 95.11 & 95.54 \\
\bottomrule
\end{tabular}
\end{table}

%

\subsection{Generalization Study}

We conduct a generalization experiment by transferring from the outdoor KITTI dataset to another outdoor dataset, ETH-Challenging~\cite{Chall_3d}. Note that the KITTI and ETH-Challenging datasets use Velodyne-64 3D LiDAR and Hokuyo 2D LiDAR sensors, respectively, resulting in significantly different appearances and distributions of point clouds. Hence, this generalization study is both practical for real-world applications and effective in demonstrating the robustness of various methods.

Tab.~\ref{tab:generalization_results} also presents RRE, RTE, and RR. Our method achieves satisfactory accuracy and robustness, exhibiting superior generalization performance compared to the coarse-to-fine baseline GeoTransformer and other point-wise descriptors. Notably, all point-wise methods show lower registration recall in the generalization setting compared to the patch-wise local descriptor SpinNet~\cite{ao2021spinnet}. This difference arises primarily because SpinNet employs a feature pyramid network architecture to learn features with abundant global context, which benefits generalization.

Furthermore, we perform an unsupervised domain adaptation (UDA) experiment on MT-PCR, tuning the network to align a point cloud to itself after random rotation and cropping. The results indicate that our model can easily adapt to an unseen domain and achieve robust and accurate performance after only one epoch of unsupervised tuning (approximately 20 minutes on an NVIDIA RTX 3090 GPU).

\begin{table}[ht]
    \centering
    \caption{Generalization results on KITTI $\to$ ETH datasets.}
    \label{tab:generalization_results}
    \begin{tabular}{l|ccc}
        \toprule
Method & RTE (cm) & RRE (°) & RR (\%) \\
        \midrule
        FCGF            & 9.08              & 0.94             & 45.86            \\
        Predator        & 11.72             & 1.38             & 65.64            \\
        SpinNet         & 6.05              & 0.98             & 99.44       \\
        GeoTransformer  & 5.97              & 0.73             & 91.87            \\
        CAST            & 6.86              & 0.66             & 97.05            \\
        MT-PCR          & 6.67              & 0.67             & 98.88        \\
        MT-PCR + UDA    & \textbf{4.27}     & \textbf{0.53}    & \textbf{99.86} \\
        \bottomrule
    \end{tabular}
\end{table}

\section{Conclusion}

In this paper, we introduced MT-PCR, the first hierarchical point cloud registration framework that unifies the strengths of Mamba and Transformer architectures. By leveraging the linear complexity and long-range modeling capabilities of Mamba, along with the spatially aware refinement of Transformers, our framework addresses the scalability limitations of existing Transformer-based PCR methods. To bridge the gap between sequence-based models and unordered 3D point clouds, we proposed a feature-serialization approach based on Z-order space-filling curves, which preserves spatial locality and improves the compatibility of point-cloud features with Mamba. Additionally, we observed that removing conventional order-indicator tokens, commonly used in sequential modeling, leads to superior registration performance, highlighting a useful design insight for applying state space models to geometric data. Comprehensive experiments on challenging benchmarks such as 3DMatch, 3DLoMatch, KITTI, and the ETH benchmarks demonstrate that \ours~not only achieves SoTA accuracy but also significantly improves computational efficiency by reducing both GPU memory consumption and FLOPs. We believe that the hybrid design principles of MT-PCR open up new opportunities for efficient and scalable 3D perception and provide a promising direction for future research in point cloud representation learning and registration.


{\small
\IfFileExists{cvmpaper_finalcopy.bbl}{\input{cvmpaper_finalcopy.bbl}}{\bibliographystyle{cvm}\bibliography{cvmbib}}
}

\end{document}